\documentclass[journal]{IEEEtran}
\IEEEoverridecommandlockouts
\usepackage[numbers, sort&compress]{natbib}

\usepackage{amsmath,amssymb,amsfonts, bm}
\usepackage[ruled]{algorithm2e}
\usepackage{subfigure}
\usepackage{graphicx}
\usepackage{epstopdf}
\usepackage{textcomp}
\usepackage{xcolor}
\usepackage{authblk}
\usepackage[marginal]{footmisc}
\usepackage{multirow}
\usepackage{multicol}
\usepackage{makecell}
\usepackage{pifont}
\usepackage{bbding}

\def\BibTeX{{\rm B\kern-.05em{\sc i\kern-.025em b}\kern-.08em
    T\kern-.1667em\lower.7ex\hbox{E}\kern-.125emX}}

\title{TiBGL: Template-induced Brain Graph Learning for Functional Neuroimaging Analysis }
\author{Xiangzhu Meng$^*$, Wei Wei, Qiang Liu, IEEE, Member, Shu Wu, IEEE, Senior Member, Liang Wang, IEEE, Fellow \thanks{Xiangzhu Meng and Wei Wei are corresponding authors (denoted by $\ast$)} \thanks{X. Meng is with the Center for Research on Intelligent Perception and Computing, Institute of Automation, Chinese Academy of Sciences, Beijing, China (xiangzhu.meng@cripac.ia.ac.cn)} \thanks{W. Wei is with Center for Energy, Environment \& Economy Research, School of Management, Zhengzhou University (weiwei123@zzu.edu.cn)} \thanks{Q. Liu, W. Shu and L. Wang are with Center for Research on Intelligent Perception and Computing Institute of Automation, Chinese Academy of Sciences, and School of Artificial Intelligence, University of Chinese Academy of Sciences (qiang.liu, shu.wu, liang.wang\}@nlpr.ia.ac.cn)} }

\begin{document}

\maketitle

\begin{abstract}
In recent years, functional magnetic resonance imaging has emerged as a powerful tool for investigating the human brain's functional connectivity networks. Related studies demonstrate that functional connectivity networks in the human brain can help to improve the efficiency of diagnosing neurological disorders. However, there still exist two challenges that limit the progress of functional neuroimaging. Firstly, there exists an abundance of noise and redundant information in functional connectivity data, resulting in poor performance. Secondly, existing brain network models have tended to prioritize either classification performance or the interpretation of neuroscience findings behind the learned models. To deal with these challenges, this paper proposes a novel brain graph learning framework called Template-induced Brain Graph Learning (TiBGL), which has both discriminative and interpretable abilities. Motivated by the related medical findings on functional connectivites, TiBGL proposes template-induced brain graph learning to extract template brain graphs for all groups. The template graph can be regarded as an augmentation process on brain networks that removes noise information and highlights important connectivity patterns. To simultaneously support the tasks of discrimination and interpretation, TiBGL further develops template-induced convolutional neural network and template-induced brain interpretation analysis. Especially, the former fuses rich information from brain graphs and template brain graphs for brain disorder tasks, and the latter can provide insightful connectivity patterns related to brain disorders based on template brain graphs. Experimental results on three real-world datasets show that the proposed TiBGL can achieve superior performance compared with nine state-of-the-art methods and keep coherent with neuroscience findings in recent literatures.
\end{abstract}


\section{Introduction}
\label{sec:intro}
With the widespread application of modern medical imaging technologies, magnetic resonance imaging (MRI) \cite{terreno2010challenges} technologies have proven to be a valuable tool for investigating neuroscience-related issues, particularly in the diagnosis of neurological disorders. In particular, functional magnetic resonance imaging (fMRI) \cite{bennett2010reliable,glover2011overview} is one of the non-invasive techniques to observe the temporal dynamics of blood oxygen level dependency (BOLD) response. In recent years, fMRI data has been widely utilized to gain a better understanding of the functional activities and organization of the human brain. For instance,  functional organization can be usually characterized by the synchronization of fMRI time series among brain regions \cite{fornito2015connectomics}. Recent researches \cite{thiebaut2022emergent, lee2022solving} suggest that functional connectivities among regions of interest (ROIs) within the brain play a crucial role in influencing behavior, cognition, and brain dysfunction. To analyze those meaningful connectivity patterns, machine learning \cite{larranaga2006machine} based works have been widely utilized in functional neuroimaging scenarios, such as  disease diagnosis \cite{khosla2019machine, pervaiz2020optimising}, individual demographic information (i.e., intelligent quotient and gender) \cite{douw2014healthy, gong2011brain}, cognitive ability \cite{sui2020neuroimaging,uddin2021brain}, etc.

As the most representative machine learning technology, deep learning \cite{li2021survey, zhou2020graph, salehinejad2017recent} has been widely investigated to learn spatial, temporal, and connective patterns of fMRI time series for diagnosing brain disorders. For examples, BrainNetCNN \cite{kawahara2017brainnetcnn} leverages the topological locality of brain networks to predict cognitive and motor developmental outcome scores for infants born preterm; BrainGNN designs novel ROI-aware graph convolutional layers that leverage the topological and functional information of fMRI. With wide applications of transformers \cite{vaswani2017attention,tay2022efficient}, there also exist several transform-based works for human brain, such as BRAINNETTF \cite{kan2022brain}, which leverages the unique properties of brain network data to maximize the power of transformer-based models for brain network analysis. Notably, how to interpret the neuroscience findings behind learnt models is as important as classification performance for functional neuroimaging analysis. Due to the black box issue of deep learning methods, these works usually cannot provide explicit interpretation to understand why a certain prediction is made. To address this issue, these works \cite{qiu2020development, li2021braingnn, zhu2022interpretable, cui2022interpretable} propose neurological biomarkers to understand the pathological mechanism of brain disorders. For example, the work \cite{cui2022interpretable} proposes a globally shared explanation generator to highlight disorder-specific biomarkers including salient ROIs and important connections. Such biomarkers heavily depend on the trained deep models, which limits their robustness. Moreover, when facing limited brain network data with noise and redundant information, how to directly train a stale deep model is still uncertain and full of challenges. 

Different from deep learning methods, traditional machine learning methods are more suitable for the case of limited medical data, so brain network models based on those technologies have been widely investigated for brain diagnosis in recent years. Specifically, to well diagnose brain disease, the neuroimage data of human brain can be firstly preprocessed to generate digitized representations, such as connectivity network among brain regions of interest (ROI), and then traditional machine learning \cite{larranaga2006machine} methods can be used to build the corresponding diagnosis model. Among these works, shadow learning-based methods firstly learn the embedding of human brain, then use classical classifier \cite{foster2014machine} to partition the learnt data into corresponding groups. Moreover, graph-based works \cite{he2010graph, hadley2016change} attempt to model the brain as a graph, allowing to tackle interesting neuroscience research questions by investigating the topological structure of brain networks.  To provide the interpretable diagnosis results, there exist interpretable works based on sparsity theory, subgraph methods. More specifically, sparsity-based works \cite{yang2015fused, coloigner2017graph,cai2018capturing} have been widely used in the diagnosis of brain disease by selecting informative features to provide interpretability for disease diagnosis. Similar to sparsity-based theory, subgraph-based works \cite{cao2015identification, lanciano2020explainable, wang2021learning} investigate to search the set of brain ROI nodes to discover patterns in the corresponding connectomes, to explain the pathology of brain disease. Most existing methods only focus on the influential ROI nodes and might neglect the important connectivity relationship in the whole brain network, which cannot additionally provide more precise explainable results.

\subsection{Motivations}

Even though above existing methods have obtained promising performance for brain disorder diagnosis in some certain situations, there still remain two important aspects that have yet to be thoroughly investigated comprehensively.
\begin{itemize}
    \item[1.] Due to expensive costs and environmental diversity of data acquisition, limited brain data with accurate labels and clean features might result in erratic results of deep learning methods. Meanwhile, without guided knowledge of neurological disorders, brain graph models might be further disturbed by the noise and redundant information in original brain networks. 
    
    \item[2.] For the issue of brain disease diagnosis, learning interpretable models is as important as mere classification performance. However, most of existing brain network models either focus on the classification performance or interpret the neuroscience findings behind learnt models. Thus, how to provide a robust brain graph model with both discriminative and interpretable abilities is very necessary and meaningful.

\end{itemize}

\subsection{Contributions}
This paper proposes a novel multi-stage functional neuroimaging analysis model with both discrimination and interpretation, called Template-induced Brain Graph Learning (TiBGL). To solve the unstable shortcoming of deep models for limited brain networks, TiBGL proposes template-based brain graph learning to capture template graphs from two intra-group and inter-group aspects, which is motivated by medical research findings \cite{satterthwaite2015linked, sporns2022structure}. The learned template graphs not only highlight those important connectivity patterns in each group but also remove the noise and redundant information in brain networks. Then, template-induced convolutional neural network is developed to fuse rich information from brain networks and learned template graphs for brain disorder diagnosis, which can generate better subject-level brain networks. Furthermore, TiBGL utilizes template graphs as augmented group-level brain network to explore those meaningful connectivity patterns related to brain disorders, providing an insightful brain interpretation analysis. We evaluate the effectiveness of the proposed TiBGL on two neurological disorders diagnoses, including Autism Spectrum Disorder (ASD) and Attention Deficit Hyperactivity Disorder (ADHD), and one brain classification task of gender identification. The major contributions of this paper are summarized as follows:
\begin{itemize}
    \item[1.] TiBGL proposes template brain graph learning based on recent neuroscience findings to extract meaningful template graphs to enhance the representative capabilities of group-level and subject-level brain networks.

    \item[2.] TiBGL utilizes template brain graphs as augmentation means for brain networks, and proposes template-induced convolutional network and brain interpretation analysis for achieving discriminative and explainable disease diagnosis tasks.
    
    \item[3.] The experimental results on three tasks of brain network classification can demonstrate that the proposed method outperforms its counterparts and provides meaningful insights for neurological disorders.
    
\end{itemize}

\subsection{Orgnization}
The remainder of this paper is organized as follows: in Section \uppercase\expandafter{\romannumeral2}, we briefly review related works on brain disorders, such as diagnosis model and interpretation analysis; in Section \uppercase\expandafter{\romannumeral3}, we describe the details of the proposed TaGBL and its optimization; in Section \uppercase\expandafter{\romannumeral4}, we conduct extensive experiments on 3 datasets to evaluate the effectiveness and robustness of our proposed TiBGL; in Section \uppercase\expandafter{\romannumeral5}, we make the conclusion of this paper.

\section{Related Works}
In the past decades, brain graph models based on functional neuroimaging analysis have been widely studied. Based on applicational targets, we roughly divide the existing methods into two categories, including diagnosis models and interpretation analysis for brain disorders.

\subsection{Diagnosis Models for Brain Disorders}
Traditional machine learning \cite{larranaga2006machine} methods have been widely used to build the corresponding diagnosis model. Among these works, shadow learning-based methods usually follow two stages. Brain networks are firstly processed to generate the embedding of human brain, then use a classical classifier to partition the learnt data into corresponding groups. Additionally, due to the high dimensionality of fMRI data, dimensionality reduction technologies should be used to reduce the dimensionality of brain features. For instance, graph-based works \cite{he2010graph, hadley2016change} attempt to model the brain as a graph, allowing it to tackle interesting neuroscience research questions by investigating the topological structure of brain networks. For these methods, if brain features from the first stage are not reliable, significant errors can be induced in the second stage. Apart from the above traditional works, deep learning methods for brain research are mainly based on Convolution Neural Networks (CNN) \cite{li2021survey, kiranyaz20211d}, Recurrent Neural Networks (RNN) \cite{salehinejad2017recent} and Graph Neural Networks (GNN) \cite{zhou2020graph}, to learn spatial, temporal and connective patterns of fMRI time series for the diagnosis of brain disorder. For instance, BrainNetCNN \cite{kawahara2017brainnetcnn} leverages the topological locality of brain networks to predict cognitive and motor developmental outcome scores for infants born preterm; BrainGNN \cite{li2021braingnn} designs novel RoI-aware graph convolutional layers that leverage the topological and functional information of fMRI. Moreover, transformers \cite{vaswani2017attention,tay2022efficient} have been studied over different types of data, and there also exist several transform-based works for human brain, such as BRAINNETTF \cite{kan2022brain}, which leverages the unique properties of brain network data to maximize the power of transformer-based models for brain network analysis. Unfortunately, limited brain data might result in erratic results of the above diagnosis models due to expensive costs of data acquisition as well as noise information in original brain networks.

\subsection{Interpretation Analysis for Brain Disorders}

In most applications of drain disorders, understanding the general pattern or mechanism associated with a cognitive task or disease is very necessary. Group-level neural findings usually highlight consistent explanations across subjects \cite{adeli2020deep, venkataraman2016bayesian, salman2019group}, such as those key ROIs as well as their connectivity. For example, class activation mapping has been used to identify salient brain regions \cite{arslan2018graph}, and to visualize effective features by gradient sensitivity \cite{yang2019interpretable}. The neurological biomarkers-based work \cite{qiu2020development, li2021braingnn,zhu2022interpretable} can be also used to provide interpretability for group-level differences.  Besides, personalized treatments \cite{brennan2019use, beykikhoshk2020deeptriage} for outcome prediction or disease sub-type detection require learning the individual-level biomarkers to achieve the best predictive performance. However, the above biomarkers heavily depend on the performance of trained deep models. When facing limited brain network data with noise and redundant information, it's full of difficulties to directly train a stale deep model with powerful ability. Different from biomarker-based works, sparsity-based works \cite{yang2015fused, coloigner2017graph,cai2018capturing} have been widely used in the diagnosis of brain disease by selecting informative features to provide interpretability for disease diagnosis. Similar to sparsity-based theory, subgraph-based works \cite{cao2015identification, lanciano2020explainable, wang2021learning} investigate to search the set of brain ROI nodes to discover patterns in the corresponding connectomes, to explain the pathology of brain disease. For example, the work \cite{lanciano2020explainable} proposes a novel approach for classifying brain networks based on extracting contrast subgraphs, i.e. a set of vertices whose induced subgraphs are dense in one class of graphs and sparse in the other, which confirms the interestingness of the discovered patterns to match background knowledge in the neuroscience literature. However, most existing methods only focus on the influential ROI nodes and might neglect the important connectivity relationship in the whole brain network, which cannot additionally provide more precise explainable results.

\section{The Proposed Method}

\begin{figure*}
\centering
  \subfigure[Original fMRI image.]{
    \centering
    \includegraphics[width=0.7\textwidth]{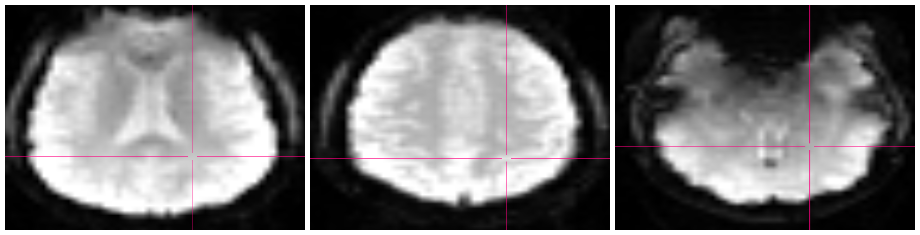}
    \label{fig:fmri}
  }
   
\subfigure[Template image of atlas.]{
    \centering
    \includegraphics[width=0.7\textwidth]{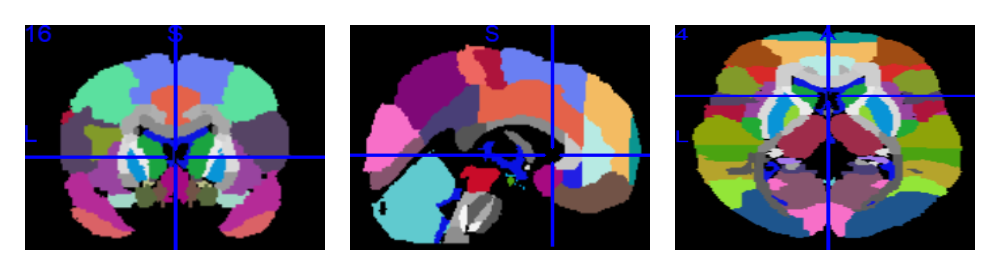}
    \label{fig:template}
  }
  
\subfigure[Time series of ROIs as well as their connectivities.]{
    \centering
    \includegraphics[width=0.35\textwidth]{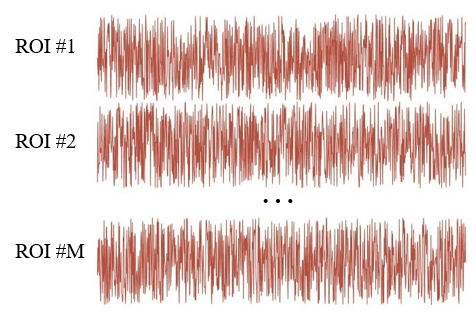}
    \includegraphics[width=0.3\textwidth]{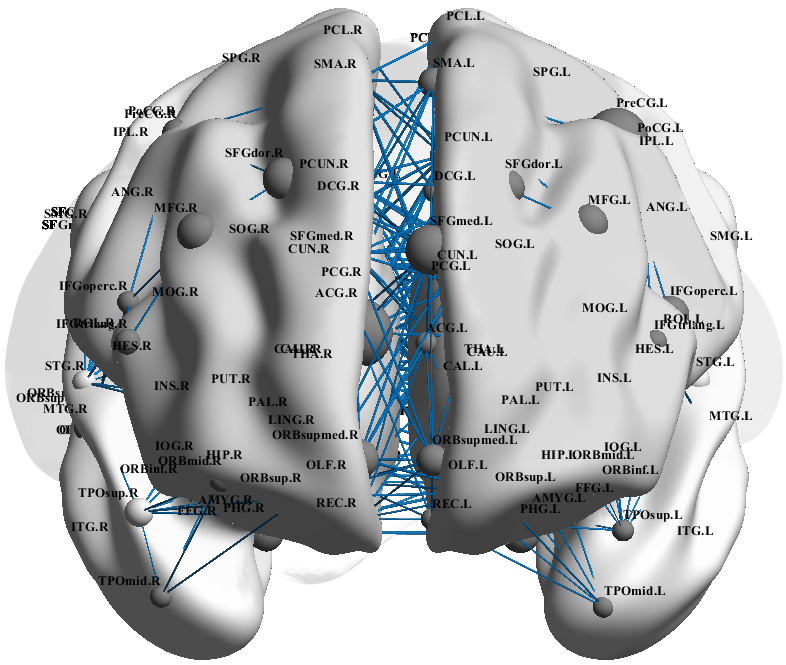}
    \label{fig:brainnet}
  }

\caption{Data processing for fMRI image.} 
\label{fig:example} 
\end{figure*} 

\begin{figure*}[htbp]
  \centering
  \includegraphics[width = \textwidth]{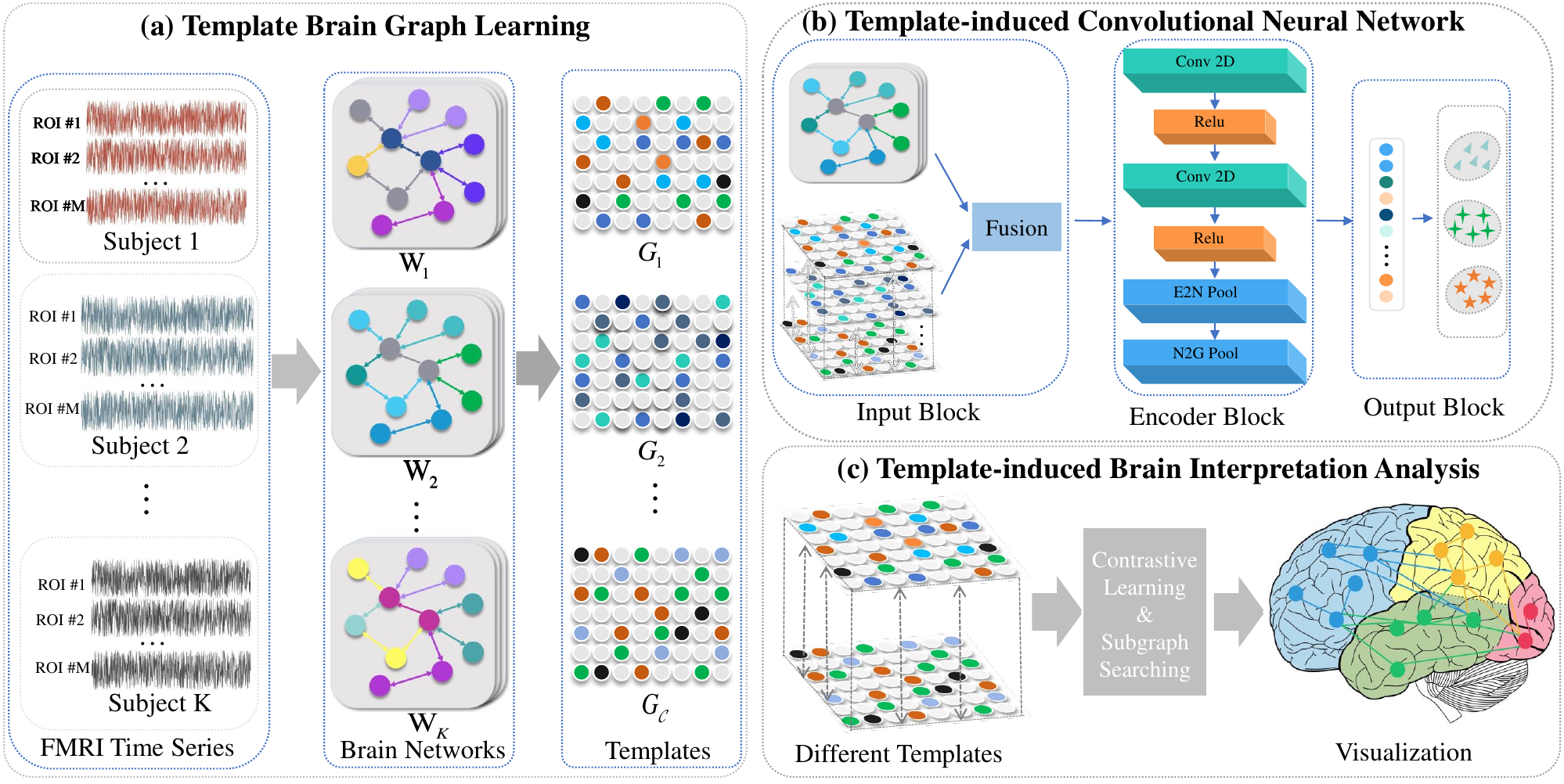}
   \caption{Flowchart of TiBGL framework, consisting of three following parts. The first part is on template brain graph learning, which extracts the meaningful template graph from all groups based on intra-group and inter-group loss functions. The second part is template-induced convolutional neural network to classify augmented brain networks with template graphs into the correct group, consisting of input block, encoder block, and output block. The third part is on template-induced brain interpretation analysis for disorder diagnosis.}
      \label{flow-chart}
\end{figure*}

In this section, we first introduce the data processing of brain networks and important notations in this paper. Then, we propose the template brain graph learning to extract meaningful template graphs for brain classification and explanation analysis. Subsequently, we propose the template-induced convolutional neural network for classification tasks. Finally, we propose template-induced brain interpretation analysis to identify connection patterns
that might be associated with specific brain disorders.

\subsection{Data Processing \& Notations}
Given one human brain neuroimage dataset that consists of $K$ resting-state fMRI scans from $\bm{\mathcal{C}}$ groups. For each resting-state fMRI scan ( as shown in Fig. (\ref{fig:fmri})), we firstly utilize the standard preprocessing procedure to preprocess original data followed by the work \cite{cui2022braingb}, including slice-timing correction, realignment, co-registration, normalization and smoothing. Then, we utilize the atalas with template image to parcellate all resting-state fMRI scans to extract $M$ mean time series of BOLD signal, where $M$ denotes the number of brain regions. As shown in Fig. (\ref{fig:template})), we visualize each brain ROIs corresponding to one specific color. Finally, we construct  functional connectivity (FC) matrix $\bm{W}_k$ as brain graph for the $k$th subject by calculating the Pearson coefficient between ROIs, as shown in Fig. (\ref{fig:brainnet}).

\subsection{Overall TiBGL Framework}
The goal of this paper is to mine these template brain graphs from original brain graph set $\{\bm{W}_k, 1 \le k \le K \}$ to simultaneously implement accurate diagnosis and explanation exploration for brain disorders. As shown in Fig. \ref{flow-chart}, the proposed TiBGL is a whole framework for functional neuroimaging analysis, consisting of three following parts. The first part is on template brain graph learning, which extracts the meaningful template graph $\{\bm{G}_c, 1 \le c \le \mathcal{C} \}$ for all groups, by minimizing intra-group and inter-group loss functions. The second part is template-induced convolutional neural network to classify augmented brain networks with template graphs into the correct group, consisting of input block, encoder block, and output block. The third part is on template-induced brain interpretation analysis for disorder diagnosis, which employs contrastive strategy to search the subgraph that reflects the functional difference between different groups. 

\subsection{Template Brain Graph Learning}
\label{sec:template}
 In this section, we attempt to extract the meaningful template graph $\{\bm{G}_c, 1 \le c \le \mathcal{C} \}$ for all groups to augment instance-level and group-level brain networks for diagnosing and explaining the neurological disorders. Inspired by two research findings \cite{sporns2022structure, satterthwaite2015linked}, we decide to exploit template graphs from two intra-group and inter-group aspects. One finding \cite{sporns2022structure} is that there are consistent patterns in functional connectivity among individuals from same group. The other finding \cite{satterthwaite2015linked} is that there are edge-level differences in functional brain connectivity matrices among different groups. Therefore, we propose intra-group and inter-group loss functions to extract the template graph $\bm{G}_c$, as shown in Fig. \ref{flow-chart}(a).

\textbf{Intra-group Template Brain Graph Learning.}
Inspired by the first finding, we can regard the template graph $\bm{G}_c$ as the latent consistent pattern in the $c$th group. For this reason, we need to minimize the difference between template graph and brain graph matrices within the same group. At the same time, we consider that subjects might play roles of different importance in learning template graph $\bm{G}_c$. Inspired by the adaptively weighting strategy \cite{nie2017auto, meng2021multi, meng2022unified}, we can adaptively allocate suitable weight for each subject in learning template graphs. To this end, the above considerations can be formulated as 
\begin{equation}
\label{adpt:1}
\begin{split}
& \bm{L}_{intra} = \sum_{c=1}^{\mathcal{C}}\sum\limits_{k\in \mathcal{I}_c} { \bm{\alpha}(c,k) \left\| \bm{G}_c - \bm{W}_k \right\|}_{_F}^2, \\
\end{split}
\end{equation}
where $\mathcal{I}_c$ denotes the set of the indexes of samples in $c$th group. $\bm{\alpha}(c,k)=1/{\left\| \bm{G}_c - \bm{W}_k \right\|}_{_F}^{\frac{1}{2}}$ is the adaptively allocated weight for the $k$th subject in learning the template $\bm{G}_c$, which refers to the definition in the work \cite{nie2017auto}. Differing from considering equal importance for all subjects, we can mine more refined information.

\textbf{Inter-group Template Brain Graph Learning.}
To increase the discriminative ability of template graphs, we attempt to maintain a large (finite) distance between elements in different template graphs. Inspired by the second finding, we attempt to maintain a margin of safety around the edge-level boundaries. That is to say, we require any two template graphs $\bm{G}_{c_1}$ and $\bm{G}_{c_2}$ to must satisfy the following inequality condition
\begin{equation}
\small
\begin{split}
& {| \bm{G}_{c_1}(i,j)-\bm{G}_{c_2}(i,j) |} \ge \gamma,  \\
\end{split}
\end{equation}
where $\gamma \ge 0$ is a user-defined parameter. In other words, one template graph is different from another template graph that invades the perimeter plus $\gamma$ margin defined by the edge-wise difference between two template graphs. Combining the above equation with margin strategy \cite{2009Distance}, we propose the inter-group loss function that  enlarges the discrepancy between different templates, which can be formulated as
\begin{equation}
\small
\label{eq:contrat_graph}
\begin{split}
& \bm{L}_{inter} = \sum_{c_1 \neq c_2}\sum_{i=1}^M\sum_{j=1}^M{[ | \bm{G}_{c_1}(i,j)-\bm{G}_{c_2}(i,j) | - \gamma ]}_{+}, \\
\end{split}
\end{equation}
where the term ${[z]}_{+}=max(z,0)$ denotes the standard hinge function.

\textbf{Overall framework.} We combine the aforementioned two components as well as sparsity normalization term into one framework, and the final objective function can be formed as follows:

\begin{equation}
\small 
\label{eq:overall}
\begin{split}
& \bm{L}_{Group} =  \underbrace{\sum_{c=1}^{\mathcal{C}}\sum\limits_{k\in \mathcal{I}_c} { \bm{\alpha}(c,k) \left\| \bm{G}_c - \bm{W}_k \right\|}_{_F}^2}_{Intra-group \ Loss} + \lambda_1 \underbrace{\sum_{c=1}^{\mathcal{C}}{\left| \bm{G}_c \right|}_1}_{Sparsity \ Term}\\
& \quad + \lambda_2 \underbrace{\sum_{c1 \neq c2}\sum_{i=1}^M\sum_{j=1}^M{[ | \bm{G}_{c_1}(i,j)-\bm{G}_{c_2}(i,j) | - \gamma ]}_{+} }_{Inter-group \ Loss},
\\
\end{split}
\end{equation}
where $\lambda_1$ and $\lambda_2$ are two hyper-parameter to balance above loss terms. Differing from those globally shared mask-based methods \cite{wang2021modeling, cui2022interpretable, kan2022fbnetgen}, learned template graphs $\bm{G}_c$ can prompt consistence information in intra-group networks and the edge-level difference between inter-group brain networks. In this way, we can utilize these template graphs as the  basic component to help the following tasks of brain disorder diagnosis as well as its interpretation analysis.

\textbf{Optimization.} With the alternating optimization strategy, we can solve the variables in Eq. (\ref{eq:overall}) in turns. To be specific, with all template graphs but $\bm{G}_c$ fixed, we can obtain the following optimization problem for template graph $\bm{G}_c$, 
\begin{equation}
\small
\label{eq:optim}
\begin{split}
& \bm{L}_{Group} =  \sum\limits_{k\in \mathcal{I}_c} { \bm{\alpha}(c,k) \left\| \bm{G}_c - \bm{W}_k \right\|}_{_F}^2 + \lambda_1{\left| \bm{G}_c - \bm{W}_k \right|}_1\\
&  \qquad + \lambda_2 \sum_{c \neq c_1}\sum_{i=1}^M\sum_{j=1}^M{[ | \bm{G}_{c}(i,j)-\bm{G}_{c_1}(i,j) | - \gamma ]}_{+}, \\
\end{split}
\end{equation}
It's easy to find that the Eq. (\ref{eq:optim}) can be seen as one typical form of the least absolute shrinkage and selection operator (LASSO) \cite{tibshirani2011regression} model. Referring to the works related to LASSO, the LARS \cite{efron2004least} algorithm can be employed to solve the template graph $\bm{G}_c$ in Eq. (\ref{eq:optim}). In this way, we can get the whole optimization process for Eq. (\ref{eq:overall}). Firstly, we use the mean values in all groups to initialize all template graphs. Then, we iteratively solve the Eq. (\ref{eq:optim}) for $\bm{G}_c$ until all variables converge. The whole procedure to solve Eq.(\ref{eq:overall}) is summarized in \textbf{Algorithm \ref{algorithm}}.

\begin{algorithm}
\caption{The optimization procedure for TiBGL}
\label{algorithm}
\LinesNumbered

\KwIn{Brain graph set $\{\bm{W}_k, c_k,  1 \le k \le K \}$, the hyper-parameters $\lambda_{1}$ and $\lambda_{2}$, the number $|\mathcal{I}_c|$ of indexes of samples in $c$th group.
}

\For{c=1:$\mathcal{C}$}{

    \For{c=1:$\mathcal{C}$}{
    Initialize the allocated weight $\bm{\alpha}(c,k)= \frac{1}{|\mathcal{I}_c|}$.
    }

    Initialize template graph $\bm{G}_c=\frac{1}{|\mathcal{I}_c|}\sum\limits_{k\in \mathcal{I}_c}\bm{W}_k$.

}

\While{not converged}{
    \For{c=1:$\mathcal{C}$}{
        \For{c=1:$\mathcal{C}$}{
         Update the allocated weight $\bm{\alpha}(c,k)=1/{\left\| \bm{G}_c - \bm{W}_k \right\|}_{_F}^{\frac{1}{2}}$.
        }
        
        Use LARS to solve the template graph $\bm{G}_c$ in Eq. (\ref{eq:optim}).
    }

}

\KwOut{Template graphs $\{\bm{G}_c, 1 \le c \le \mathcal{C} \}$.}
\end{algorithm}

\subsection{Template-induced Convolutional Neural Network}
Motivated by the observation that certain types of inductive biases on spatial relations can be beneficial to classification tasks, we propose a novel template-induced convolutional neural network. As shown in Fig. \ref{flow-chart}(b), it mainly adopts three following blocks to implement the diagnoisis of brain disorders.

\textbf{Input Block.}
Given a brain graph $\bm{W}_k$, we construct multiple source representations for brain graph $\bm{W}_k$, by learnt template graphs. To be specific, we sum all group-level template graphs as global template graph $\bm{G}_c$, and then fuse global template graph $\bm{G}_c$ and brain network $\bm{W}_k$ in element-wise manner, i.e., $\bm{W}_k \odot  \bm{G}$. The above operator can be seen as the data augmentation based on group-level template graphs for brain networks. Compared to original brain networks, augmented brain graphs pay more attention to those important connectivities, by removing the irrelated and noise information, which have the more discriminative ability.

\textbf{Encoder Block.}
We design the encoder block to extract the features of brain graphs after data augmentation. To be specific, we design a convolutional neural network that mainly contains three layer types: edge-to-edge filter layer, edge-to-node pooling (E2N Pool) layer, and node-to-graph pooling (N2G Pool) layer. Edge-to-edge filter layer consists of one or more simple convolutional filters of a particular shape and performs the specific operation on the brain graph. E2N and N2G Pool layers are adopted to downsample the data from edges and nodes to obtain the graph-level representation, which refers to the work \cite{kawahara2017brainnetcnn}. It takes feature maps from the input block as input and then outputs a distinct feature map $\bm{{f}^H}(\bm{W}_k \odot  \bm{G})$ for the output block.

\textbf{Output Block \& Loss Function.}
After the processing of encoder block, we can obtain its output as the final hidden feature. Finally, we use one-layer linear network $\bm{f}_O$ as the output layer. According to the above consideration, we can obtain the output of the whole network as follows:
\begin{equation}
    \begin{split}
        \hat{\bm{y}}_k = \bm{f}_O(\bm{{f}_H}(\bm{W}_k \odot  \bm{G})),
    \end{split}
\end{equation}
where $\hat{\bm{y}}_k$ denotes the predicted group of the $k$th subject. For the classification tasks, we can employ cross entropy as the loss function to train the whole neural network. Finally, we utilize the stochastic gradient descent (SGD) method to update all weight parameters.

In this way, we combine template graphs with convolutional neural networks into an end-to-end framework for brain disorder diagnosis. Compared to other deep learning works, the main advantage is that the prior information of the template graph is beneficial for end-to-end training with limited subjects, leading to potential performance improvements.

\subsection{Template-induced Brain Interpretation Analysis}
The mere predictive power of the proposed TiBGL is of limited interest to neuroscientists, for which there exist plenty of tools for the diagnosis of specific mental disorders. What matters is the interpretation of TiGBL, as it can provide novel insights and hypotheses. However, 
most existing biomarker-based brain network works usually record the top $K$ connectivities to identify connectivity patterns that highlight those important ROIs and connections. Obviously, there exists a gap between such artificial construction manner and related findings that individuals in disease groups often exhibit weakened functional connectivity in neural networks. To overcome this limitation, we attempt to employ the learned template graphs as group-level brain networks and combine them with the related finding to identify connection patterns that might be associated with specific brain disorders.
 
 To be specific, we attempt to explore the subgraph $\bm{E}$ as connectivity patterns for the explanation analysis of brain disorders, motivated by frequent subgraph mining tasks \cite{boden2012mining, yan2008mining}. We use $M$-dimensional vector $\bm{e}$ to denote the nodes set $\bm{V}$ of subgraph $\bm{E}$. If the $i$th ROI node in the nodes set $\bm{V}$, $\bm{e}_i=1$; otherwise, $\bm{e}_i=0$. Let $\bm{E'} = {\bm{e}_i}^T \bm{e}_i$, which can be seen as a fully connected graph based on nodes set $\bm{V}$. To significantly distinguish the difference between two groups of brain graphs, we expect that the graph $\bm{E}'$ is close to one template graph $\bm{G}_{c_1}$, but far away from the other template graph $\bm{G}_{c_2}$. Based on the observation of weakened connectivity in functional domains of patients, we can formulate the following problem to search node set in subgraph $\bm{E}$:
\begin{equation}
\label{eq:sub1}
\begin{split}
&\mathop{\min} \frac{1}{2}{\left\| {\bm{G}_{c_1}}-\bm{E}' \right\|}_{_F}^2-\frac{1}{2}{\left\| {\bm{G}_{c_2}}-\bm{E}' \right\|}_{_F}^2 + \eta {\left\| \bm{E}' \right\|}_1 \\
& \ \ \ \  s.t. \quad \bm{e}_i= 0 \ or \ 1, 1 \le i \le M,\\
\end{split}
\end{equation}
where the third term is the regularization term penalizing solutions, and $\eta>0$ is the hyper-parameter. For the convenience of solving Eq. (\ref{eq:sub1}), Eq.(\ref{eq:sub1}) can be rewritten as the following equation by expanding the above equation and simplifying it by neglecting the constant terms:
\begin{equation}
\label{eq:sub2}
\begin{split}
&\mathop{\min} tr((\bm{G}_{c_1}-\bm{G}_{c_2})\bm{E}') + \eta {\left\| \bm{E}' \right\|}_1 \\
& \ \ \ \  s.t. \quad \bm{e}_i=0 \ or \ 1,, 1 \le i \le M.\\
\end{split}
\end{equation}
Notably, the first term also is seen as the similarity between fully connected graph $\bm{E}'$ and the difference $\bm{G}_{c_1}-\bm{G}_{c_2}$ between two template graphs. Once we get the nodes set $\bm{e}$ by solving the above equation, we can need to determine if exists the relation between selected ROIs. We select the edges between nodes based on the subtracted template graph, i.e. $|\bm{G}_{c_1}-\bm{G}_{c_2}|$. Mathematically, if $i$th node and $j$th node in $\bm{V}$ are connected with each other in subtracted template graph $|\bm{G}_{c_1}-\bm{G}_{c_2}|$, $\bm{E}(i,j)=1$; otherwise, $\bm{E}(i,j)=0$. Accordingly, we can generate a meaningful subgraph to explicitly explain the dissimilarity between different groups, based on the aspect of the functional connectivity among brain ROIs, as shown in Fig.(\ref{flow-chart})(c). Notably, Eq. (\ref{eq:sub2}) is a non-deterministic polynomial-time hardness (NP-hard) problem. To effectively solve the nodes set $\bm{e}$, we can transform the Eq. (\ref{eq:sub2}) into semidefinite programming (SDP) and then refine it by the local-search procedure, referring to the works \cite{2013Denser, lanciano2020explainable}. Compared TiBGL with other related subgraph works \cite{yang2018mining, wu2016mining} that are NP-hard and hard to approximate, it has been proved that this approximation is much better. More importantly, the subgraph $\bm{E}$ guided by guided knowledge might mine those more reasonable and refined neuroscience findings from clinical human brain data.

To this end, the proposed TiBGL can not only provide the robust brain network identification model but also explore some new neuroscience findings in clinical human brain data.

\section{Experiments and Analysis}
In this section, we conduct lots of experiments to comprehensively validate the effectiveness and explanation of the proposed TiBGL.
In this section, our goal is to answer the following questions:

\begin{itemize}
   \item RQ1. Can TiBGL perform better than other counterparts in brain classification?

   \item RQ2. Can TiBGL provide meaningfully interpretable findings for neurological disorders?

    
   \item RQ3. Why does TiBGL work in the diagnosis of neurological disorders?

    \item RQ4. What are the advantages of the proposed TiBGL?
\end{itemize}

\subsection{Datasets and Preprocessing}
\textbf{Datasets.}
To fully validate the identification performance of our proposed framework, massive experiments are performed on three tasks of brain graph classification, including two types of brain neuro-science disorder diagnosis and gender classification. More specifically, we collect the brain graph data from two publicly available and one self-organized datasets, which can be introduced as follows:
\begin{itemize}

    \item \textbf{Autism Brain Imagine Data Exchange (ABIDE)} \footnote{http://preprocessed-connectomes-project.org/abide/download.html} provides previously collected rs-fMRI ASD and matched controls data for the purpose of data sharing in the scientific community. ABIDE contains 1112 subjects, which are composed of structural and resting state fMRI data along with the corresponding phenotypic information. In particular, we focus on the portion of dataset containing adolescents, which contains individuals whose ages are between 15 and 20 years. Thus, 116 subjects can be divided into conditional group, labeled as ASD, and 121 subjects can be divided into control group, labeled as TD.

    \item \textbf{ Attention Deficit Hyperactivity Disorder (ADHD)} takes from USC Multimodal Connectivity Database (USCD)\footnote{http://umcd.humanconnectomeproject.org}, which collects rs-fmri ADHD and matched controls data from multiple sites. It's an unrestricted public release of resting-state fMRI and anatomical datasets. In particular, we select 520 subjects, which are usually used to evaluate brain graph models.  Here, we select 190 subjects in the condition group, labeled as ADHD, and 330 subjects in the control group, labeled as TD.

    \item \textbf{China-Japan Friendship Hospital (CJFH)} collects rs-fmri data for the purpose to exploit the relationship between brain functional connectivity and big five personality traits, which is one self-organized human brain dataset. CJFH contains the rs-fmri data of 346 individuals. Here, we focus on the task of gender classification. To be specific, there are 171 male individuals, labeled as Male,  and 175 female individuals, labeled as Female.
            
\end{itemize}

\textbf{Preprocessing.} For the ABIDE dataset,  we downloaded the preprocessed rs-fMRI series data from the preprocessed ABIDE dataset with Configurable Pipeline for the
Analysis of Connectomes (CPAC), band-pass filtering (0.01 - 0.1 Hz), no global signal regression, parcellatinng each brain into 116 ROIs by the Automated Anatomical Labeling (AAL) atlas.For the ADHD and CJFH datasets, we utilize the standard preprocessing procedure to process original fMRI data, as work \cite{cui2022braingb}. Then, Craddock 200 (CC200) is used to extract the brain ROIs for each subject, which parcellates each brain into 200 ROIs. The AAL atlas is employed to extract the time series of ROIs for each subject in the CJFH dataset, and we empirically neglect the brain ROIs in cerebellum for CJFH dataset, which parcellates each brain into 90 ROIs. For above datasets, we use the mean time series of ROIs to compute the correlation matrix as functional connectivity, i.e. brain graph, which provides an indicator of co-activation levels in brain regions. 

\subsection{Compared Methods and Experimental Settings}
\textbf{Compared Methods.} We evaluate the effectiveness of our framework in classifying brain graphs by comparing it with related methods. Three types of methods are chosen to compare their performance with our proposed framework. The first type of methods are traditional machine learning methods, which firstly utilize three feature transform methods, including raw features (Raw), Principle Component Analysis (PCA) \cite{wold1987principal}, and Lasso \cite{tibshirani1996regression}, and then select Support Vector Machine (SVM) as the classifier. The second type of methods are based on graph structure information,  including Graph2Vec \cite{narayanan2017Graph2Vec}, Sub2Vec \cite{adhikari2018Sub2Vec}. These works aim to transform the brain graph into the low-dimensional embedding based on the graph structure information. The third type of methods belong to deep learning methods, including Graph Convolutional Networks (GCN) \cite{wu2019simplifying}, BrainNetCNN \cite{kawahara2017brainnetcnn}, DIFFPOOL \cite{ying2018hierarchical}, and BrainGNN \cite{li2021braingnn}.

\textbf{Experimental Settings.}
 For the first type of methods, we keep the upper triangle of the matrices and flattened the triangle values to vectors, and use them as the input feature. For the second type of methods, we additionally construct the binary graph for brain graph by threshold filtering and then use it as the input. For the third type of methods, we select the original brain graph as input for BrainNetCNN and adjust suitable nodes features and graph structure for other methods with their best performances. For all datasets, we randomly select 70\% of the samples as training samples, 10\% of samples as validating samples, and the remaining 20\% of samples as testing samples at each iteration. We repeatedly run the above validation process ten times for all methods and use the accuracy of classification as the evaluation index. Finally, we summarize all experimental results to comprehensively evaluate all methods.

\begin{table*}[htbp]
\small
\caption{Comparison results (\%) of brain network classification. Black bold font denotes the best performance on the dataset, and underline denotes the second best performance on the dataset.}
\label{tab:accuracy}
\centering
\renewcommand\arraystretch{1.5}
\begin{tabular*}{0.8\textwidth}{@{\extracolsep{\fill}}c|cc|cc|cc}  
\hline
Methods & \multicolumn{2}{c|}{ABIDE} & \multicolumn{2}{c|}{ADHD} & \multicolumn{2}{c}{CJFH} \\
 & ACC & AUC & ACC & AUC  & ACC & AUC \\
\hline  
Raw & 62.08$_{\pm 3.62}$ & 67.36$_{\pm 1.35}$ & 60.69$_{\pm 5.89}$ & 53.10$_{\pm 3.89}$ & 75.29$_{\pm 4.59}$ & 77.31$_{\pm 2.76}$\\
PCA\cite{wold1987principal} & 59.72$_{\pm 5.79}$ & 65.58$_{\pm 4.62}$ & 58.61$_{\pm 5.34}$& 64.82$_{\pm 4.84}$& 73.75$_{\pm 4.92}$ & 79.80$_{\pm 5.02}$\\
Lasso\cite{tibshirani1996regression} & 59.17$_{\pm 2.50}$ & 65.78$_{\pm 1.04}$ & 59.58$_{\pm 3.43}$ & 65.54$_{\pm 5.82}$ &\underline{76.25$_{\pm 4.08}$} & \underline{80.68$_{\pm 1.89}$} \\
Graph2Vec\cite{narayanan2017Graph2Vec} & 59.03$_{\pm 3.55}$ & 63.94$_{\pm 3.00}$ & 61.54$_{\pm 2.01}$ & 69.30$_{\pm 3.31}$ & 51.44$_{\pm 5.72}$ & 53.21$_{\pm 2.77}$ \\
Sub2Vec\cite{adhikari2018Sub2Vec} & 55.56$_{\pm 5.16}$ & 62.35$_{\pm 2.28}$ & 65.51$_{\pm 1.98}$ & 70.78$_{\pm 2.09}$ & 53.27$_{\pm 3.66}$ & 55.31$_{\pm 2.98}$\\
GCN\cite{wu2019simplifying} & 57.42$_{\pm 2.41}$ & 55.56$_{\pm 2.76}$ & 64.42$_{\pm 3.95}$ & 70.19$_{\pm 1.09}$ & 68.85$_{\pm 4.06}$ & 74.05$_{\pm 3.65}$ \\
BrainNetCNN\cite{kawahara2017brainnetcnn}  & \underline{67.00$_{\pm 2.14}$} & \underline{73.02$_{\pm 5.63}$}& \underline{70.64$_{\pm 2.34}$} & \underline{77.42$_{\pm 1.59}$} & 70.19$_{\pm 3.42}$ & 74.40$_{\pm 2.58}$ \\
DIFFPOOL\cite{ying2018hierarchical} & 56.12$_{\pm 4.61}$  & 62.34$_{\pm 3.42}$  & 63.85$_{\pm 2.87}$ & 67.64$_{\pm 4.08}$ & 57.31$_{\pm 5.44}$  & 62.56$_{\pm 3.07}$ \\
BrainGNN\cite{li2021braingnn} & 63.12$_{\pm 4.61}$ & 70.11$_{\pm 4.43}$ & 65.82$_{\pm 3.78}$ & 71.03$_{\pm 1.93}$ & 70.51$_{\pm 2.42}$ & 77.93$_{\pm 3.63}$ \\
TiBGL & \textbf{71.55$_{\pm 2.87}$} & \textbf{74.96$_{\pm 2.23}$} & \textbf{76.51$_{\pm 2.42}$} & \textbf{82.05$_{\pm 4.93}$} & \textbf{80.57$_{\pm 2.14}$} & \textbf{84.67$_{\pm 1.42}$} \\
\hline
\end{tabular*}
\end{table*}

\subsection{Comparison Results and Analysis (Q1)}
To validate the superior performance of the proposed TiBGL, this paper has conducted massive experiments on the ABIDE, ADHD200, and CJFH datasets. We run all methods on the above three datasets in the same environment and then summarize all experimental validation results. To be specific, we calculate the classification accuracy (ACC), and area under curve (AUC) as the final evaluation index. As shown in Table \ref{tab:accuracy}, we summarize the mean evaluation indexes as well as their standard deviations.

For the ABIDE dataset, the proposed TiBGL gets the best performance with ACC of 71.55\% and AUC of 74.96\%. Among those methods conducted on the ABIDE dataset, TiBGL only performs 70\% on two terms of ACC and AUC. Besides, BrainNetCNN also obtains the promising performance compared to other methods, which might discover the important connectivity patterns for ASD. However, GCN and DIFFPOOL have relatively poor performance, and the main reason is that graph neural networks should be further adjusted for brain networks, such as BrainGNN.

For the ADHD dataset, the proposed TiBGL also gets the best performance, in which the performance on AUC is more than 80\%. Compared to traditional machine learning methods, deep learning-based methods have more superior performance in most situations. For example, GCN, BrainNetCNN and BrainGNN all perform 70\% on the index of AUC. Besides, Sub2Vec also obtains comparable performance with ACC of 65.51\% and AUC of 70.78\% on the ADHD dataset.  

For the CJFH dataset, this task mainly focuses on gender classification according to brain networks, which is usually used to validate the performances of brain graph models. Even though the proposed TiBGL can obtain the best performance,  deep learning-based methods have poorer performance compared to traditional machine methods. For example, Lasso have the best performance with ACC of 76.25\% and AUC of 80.68\% besides TiBGL. However, DIFFPOOL just obtain the performance with ACC of 57.31\% and AUC of 63.56\%, which is far lower than Lasso.

According to results in Table \ref{tab:accuracy} and the above discussions, we can readily find that the proposed TiBGL performs other methods in most situations. The main reason is that the template brain graphs can provide the guided knowledge to induce the classification model to pay more attention to those import brain ROIs as well as their connectivity relationship. Especially for the issue of neurological disorders diagnosis, there exists much noise and redundant information in original brain graphs, which is one key factor to limit the performance of diagnosis models. Taking ASD and ADHD as examples, it's effortless to observe that both traditional and deep learning methods cannot obtain the applicable results on the ABIDE and ADHD200 datasets, such as BrainGNN and GCN. Comparing the results of neurological disorders with gender classification task, we also find that the performances of these methods on CJFH dataset are superior to ABIDE and ADHD200 datasets, which can further show the challenges of neurological disorders. Therefore, it's necessary to first extract the prior knowledge for the subsequent neurological disorders diagnosis.

\subsection{Interpretation Analysis for Neurological Disorders (Q2)}
\label{sec:interpret}
\begin{figure}[htbp]
\includegraphics[width=0.5\textwidth]{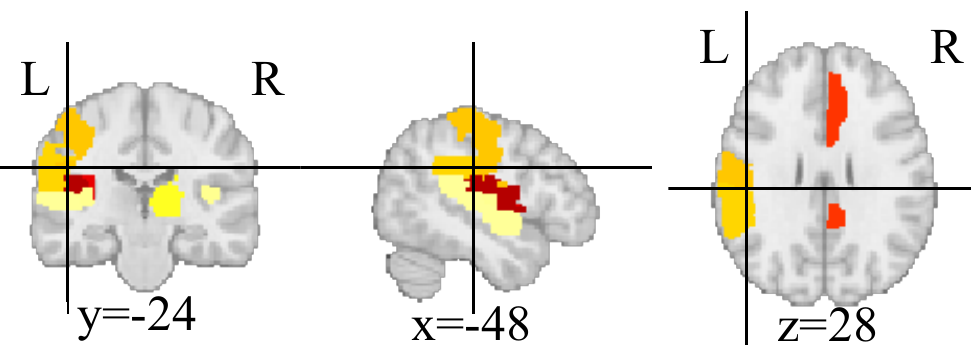}
\centering
\caption{Visualization of Brain ROIs on ABIDE dataset, which shows those important ROIs related to ASD, such as ROIs of cerebellum, prefrontal cortex, posterior parietal cortex, and middle temporal gyri, etc.}
\label{fig:rois}
\end{figure}

\begin{figure}[htbp]
\includegraphics[width=0.5\textwidth]{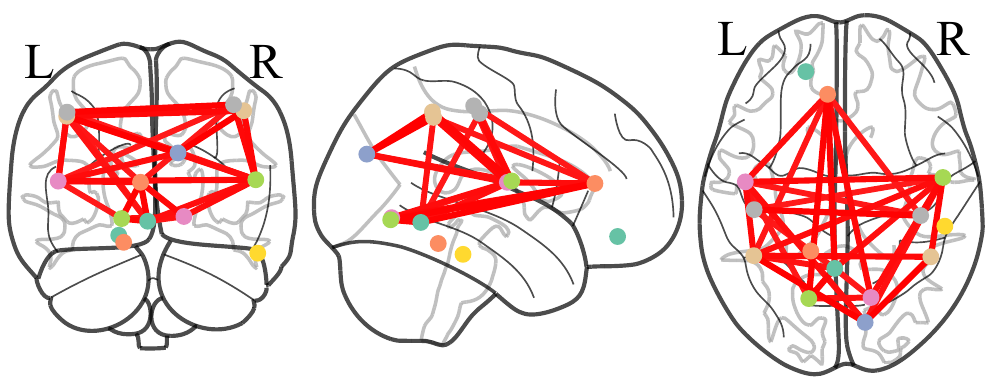}
\centering
\caption{Visualization of Subgraph on ABIDE dataset, which shows those important connectivities between brain ROIs related to ASD, such as Rolandic\_Oper\_L, Rolandic\_Oper\_R, Insula\_L, Insula\_R, Cingulum\_Mid\_L, Hippocampus\_L, Postcentral\_R, etc.}
\label{fig:connect}
\end{figure}

The accuracy of brain graph models is a predominant goal over its interpretability, which is instead a key requirement in neuroscience. To explain whether TiBGL can utilize template brain graphs to discover meaningfully interpretable findings for neurological disorders, we conduct the related validation on ABIDE dataset. Firstly, we first extract the template brain graphs $\bm{G}_{_{ASD}}$ and $\bm{G}_{_{TD}}$ for ASD and TD groups. In the later section, we additionally visualize their heatmaps in Fig.\ref{fig:finding2-1}. Then, we can obtain the critical brain ROIs that are related to ASD, by solving Eq. (\ref{eq:sub2}). To be specific, we summarize the brain ROIs as a list [ Rolandic\_Oper\_L, Rolandic\_Oper\_R, Insula\_L, Insula\_R, Cingulum\_Mid\_L, Hippocampus\_L, Postcentral\_R, SupraMarginal\_L, SupraMarginal\_R, Putamen\_L, Heschl\_L, Heschl\_R, Temporal\_Sup\_L, Temporal\_Sup\_R, Cerebelum\_9\_L, Vermis\_3 ], where each element is the abbreviation of brain ROI label. We also visualize these brain ROIs in Fig. \ref{fig:rois}. Though Fig. \ref{fig:rois}, we can find that these brain ROIs are mainly concentrated inin several regions. Furthermore, we combine the template graphs with obtained ROI sets in section \ref{sec:interpret} to generate the subgraph that contains the key ROIs as well as their connectivity. Then, we visualize the subgraph of brain network in Fig. \ref{fig:connect}. Through Fig. \ref{fig:connect}, we can find that these key connections among brain ROIs of cerebellum, prefrontal cortex, posterior parietal cortex, and middle temporal gyri, are highly related to ASD and TD classification. More importantly, the above finding on ASD is consistent with several previous studies \cite{khan2015cerebro, di2011aberrant}. This implies that the findings by TiBGL should be insightful and meaningful. To this end, the obtained subgraph of brain network can be seen as the interpretation result for neurological disorders.

\subsection{Ablation Analysis (Q3)}
To explain the reasons why the proposed TiBGL can work in the diagnosis of neurological disorders, the study of ablation analysis is conducted to evaluate the effects of template brain graphs, encoder blocks, as well as hyper-parameters. Specifically, for each test, the corresponding term is removed while retaining the other terms. 

\subsubsection{Effort of Template Brain Graph}

\begin{table}[htbp]
\small
\caption{Results (\%) of TiBGL without(with) Templates.}
\label{tab:effort_brain}
\centering
\renewcommand\arraystretch{1.5}
\begin{tabular*}{0.36\textwidth}{@{\extracolsep{\fill}}ccc}  
\hline
Templates & ABIDE & CJFH \\
\hline  
\XSolidBrush  & 67.34$_{\pm 2.14}$  & 70.19$_{\pm 4.32}$ \\
\CheckmarkBold & 71.55$_{\pm 2.87}$ & 80.57$_{\pm 2.14}$ \\
\hline
\end{tabular*}
\end{table}

To demonstrate the effort of the proposed template brain graph, we conduct experiments of TiBGL and its variant on the ABIDE and CJFH datasets. As shown in Table \ref{tab:effort_brain}, the proposed TiGBL gets the best performance on these two datasets. Compared to TiGBL without template brain graphs, our proposed TiBGL can obtain obvious improvement due to the discovered knowledge by templates. Therefore, the effect of brain network identification can further go beyond.

\begin{figure*}[htbp]
\centering
\includegraphics[width=0.35\textwidth]{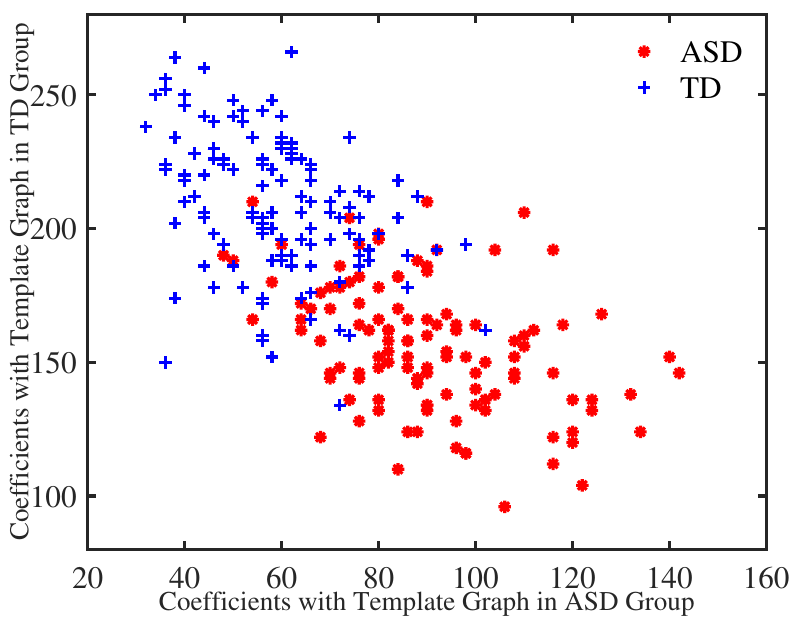}
\includegraphics[width=0.35\textwidth]{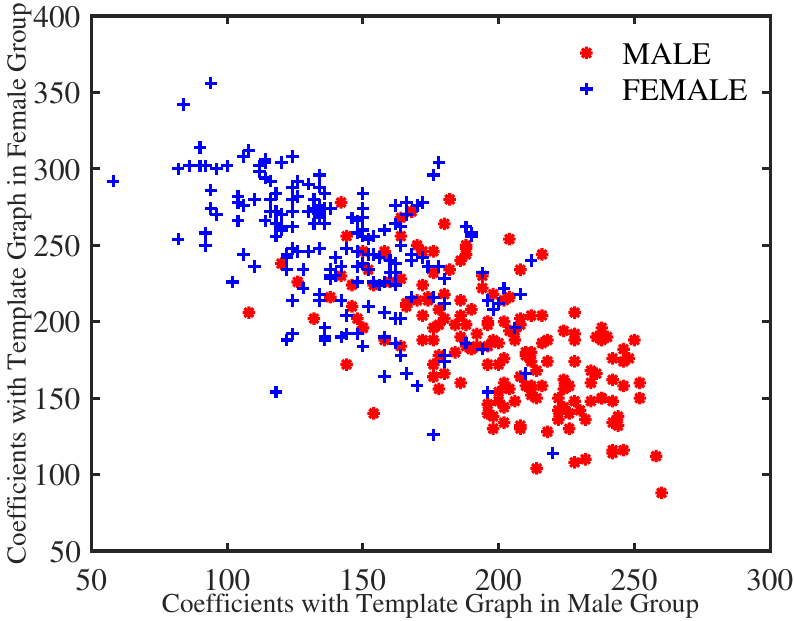}
\caption{Visualization of Similarity Scores on ABIDE and CJFH dataset. X-axis and Y-axis can stand for the coefficients with template graphs in ASD and TD groups, respectively. Visualization results demonstrate that template brain graphs can reveal the groups' edge-level differences in functional brain connectivity matrices. }
\label{fig:similiarity}
\end{figure*}

\begin{figure*}[htbp]
\centering

\includegraphics[width=0.32\textwidth]{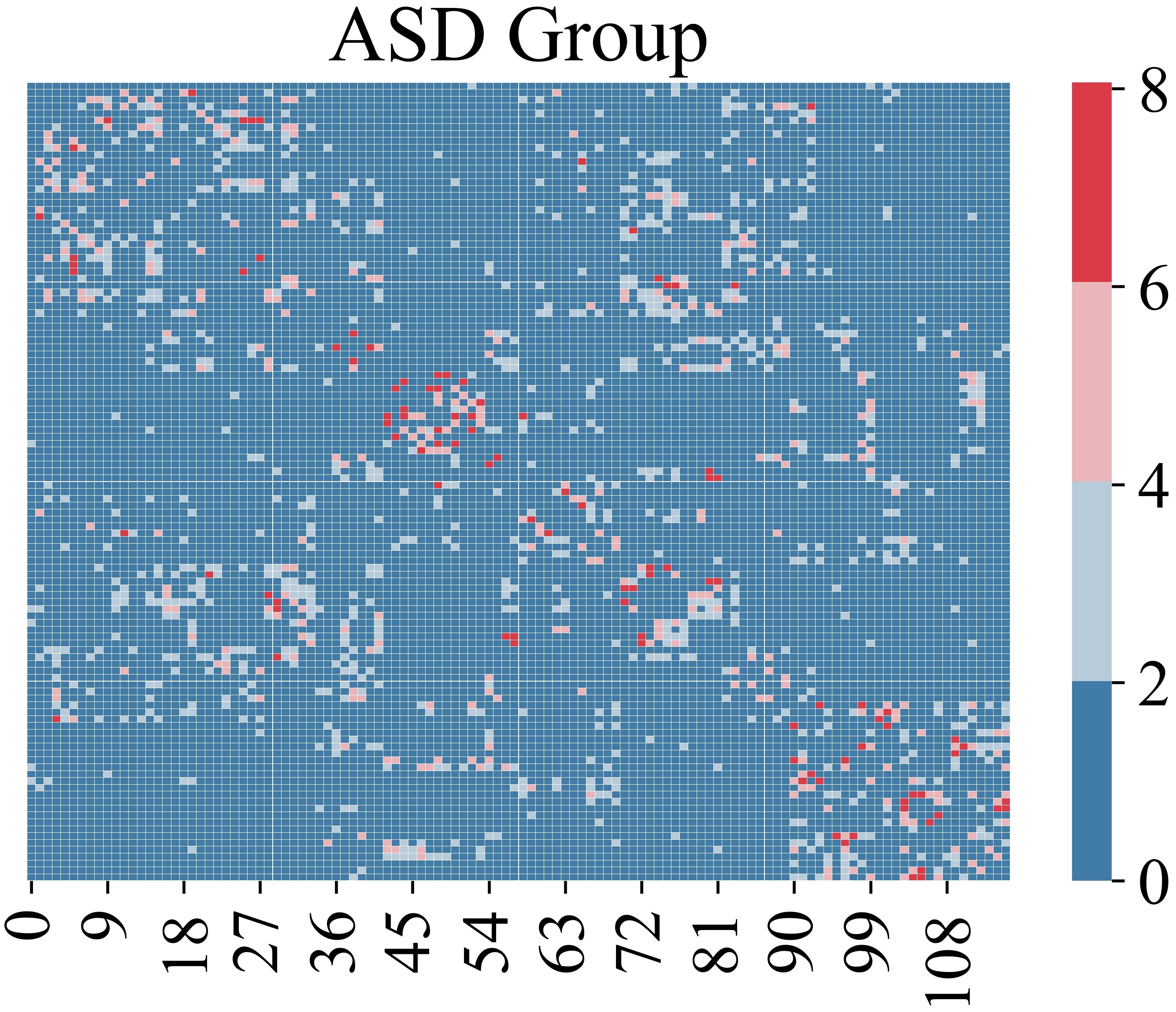}
\includegraphics[width=0.325\textwidth]{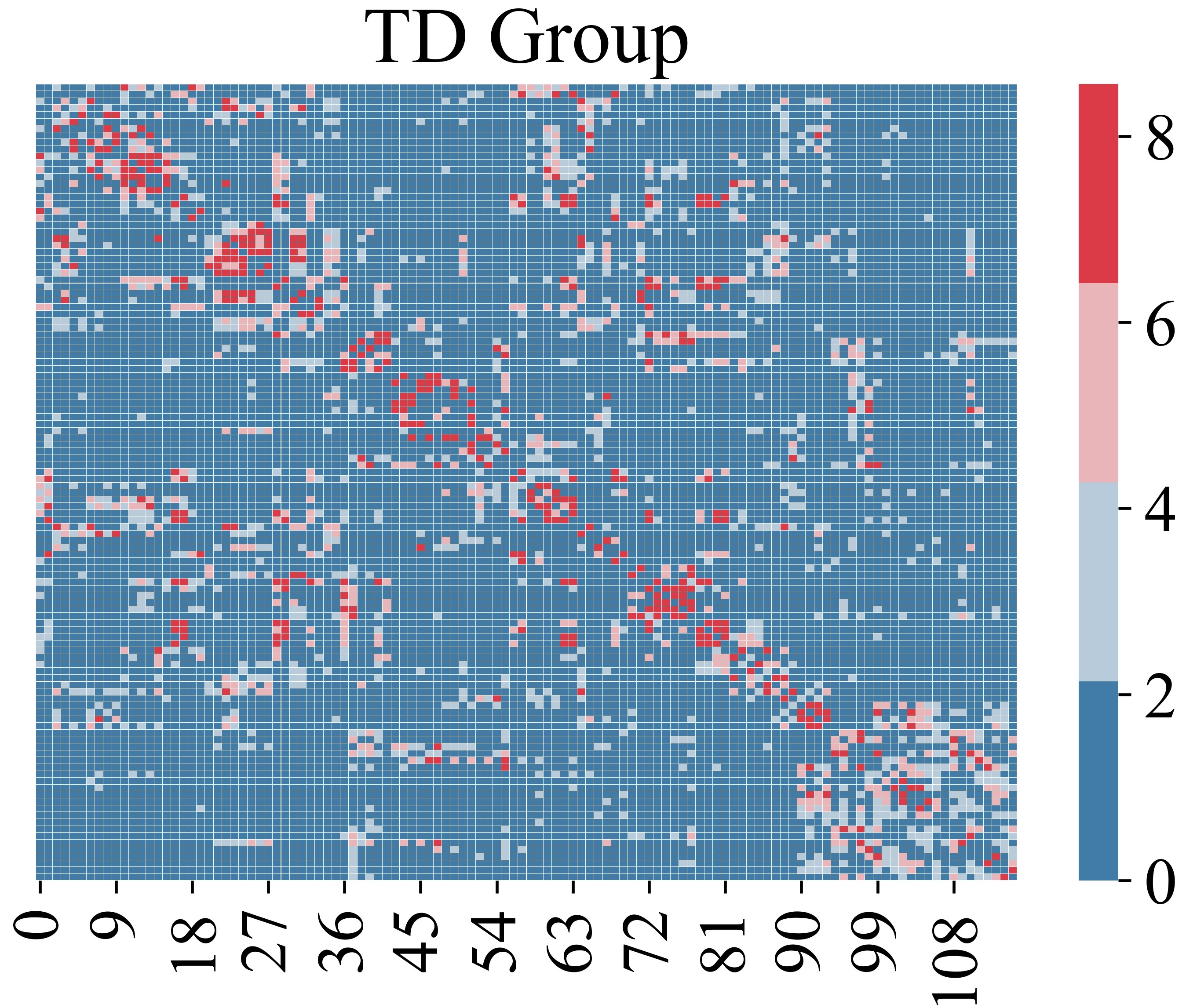}

\vspace{0.2cm}

\includegraphics[width=0.33\textwidth]{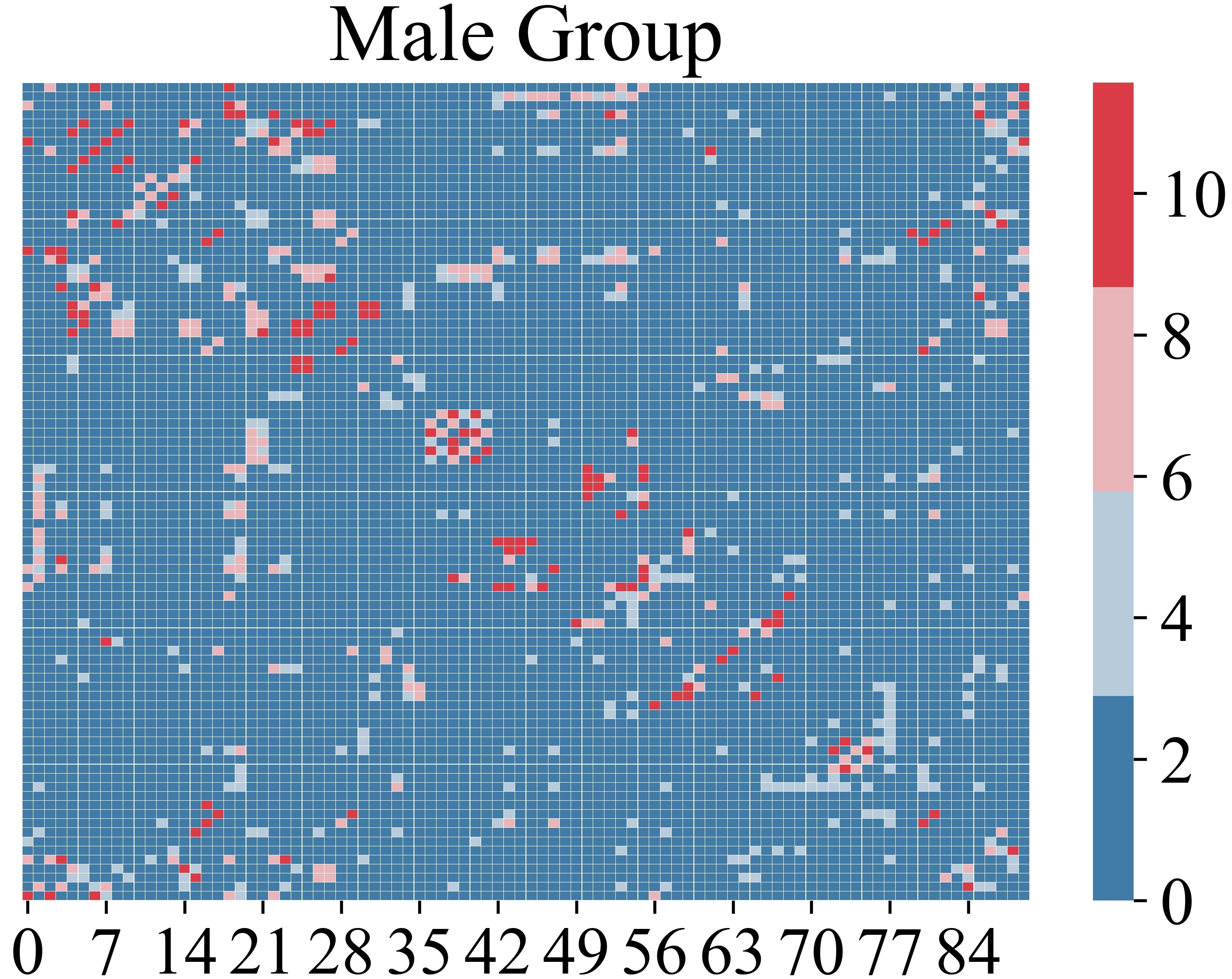}
\includegraphics[width=0.33\textwidth]{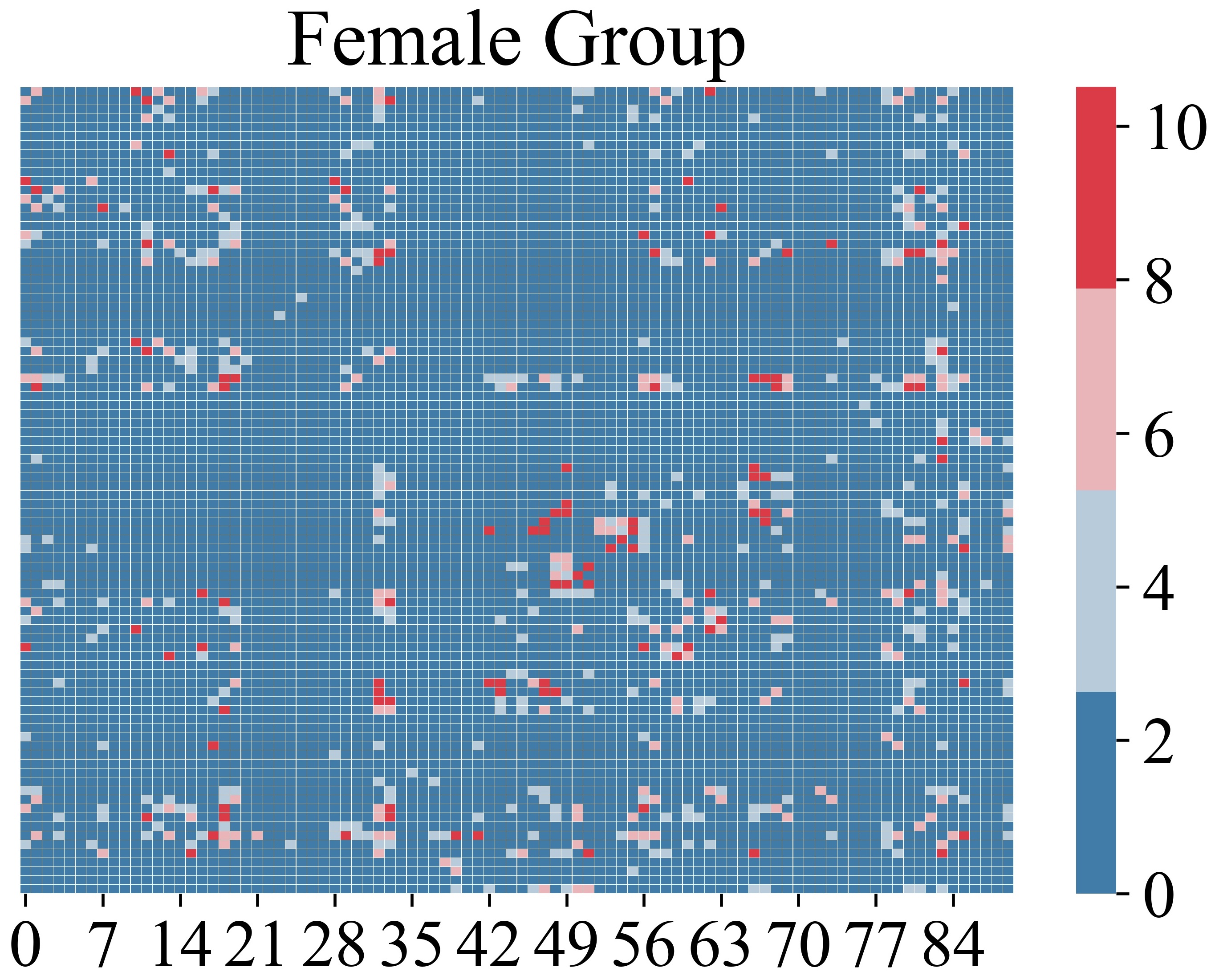}

\caption{Template Brain Graphs on ABIDE and  CJFH datasets. X-axis denotes the brain ROIs, and the colorbar can reflect the different levels of functional connectivity between brain ROIs. Visualizations of different template brain graphs in ASD and TD groups imply that template brain graphs can reveal the groups' edge-level differences in functional brain connectivity matrices. Similarly, visualizations of different template brain graphs in Male and Female groups can also validate the above observation.   }
\label{fig:finding2-1}
\end{figure*}


Besides, TiBGL extracts template graphs inspired by two research findings as illustrated in Section \ref{sec:template}. For this reason, we mainly analyze whether the obtained template graphs are consistent with these two neural findings. Firstly, we visualize the similarity scores $<\bm{W}_k,\bm{G}_c>$ in Fig. \ref{fig:similiarity}. Through the data distributions in Fig. \ref{fig:similiarity}, we can find that the similarity between given brain graph and template brain graph belonging to its group is usually larger than other template brain graphs. This implies that the template brain graph can be seen as one consistent pattern in functional connectivity among individuals in the same group, which is consistent with the first neural finding. Secondly, we summarize the heatmaps of template brain graphs in ABIDE and CJFH datasets, as shown in Fig. \ref{fig:finding2-1}. Different from dense brain graphs of individuals, the template brain graphs are more sparse, containing those edges that have important roles for its group. Sparse brain graphs can be also seen as one effective manner to eliminate noise and redundant information in original brain graphs. More importantly, there exists the structural difference between different groups, through Fig. \ref{fig:finding2-1}. This implies that template brain graphs can reveal the groups' edge-level differences in functional brain connectivity matrices, which is consistent with the neural finding \cite{satterthwaite2015linked}.

\subsubsection{Effort of Convolutional Encoder Block}
To validate the effort of convolutional blocks, we additionally propose two variants of TiBGL by substituting it with multilayer perception (MLP) and Transformer. For MLP, we flatten the fused brain graphs as vectors to predict the brain diagnose. For Transformer, we use the rows of fused brain graphs as tokens, and finally pool all tokens as global representation. To be specific, we detailly summarize the experimental result of classification accuracy on the ABIDE and CJFH datasets in Table \ref{tab:template}. According to Table \ref{tab:template}, it's obvious that TiBGL with convolutional blocks can get better performance. The main reason is that CNN with fewer parameters might be suitable such situations with limited brain graphs. Therefore, we can employ such convolutional blocks in brain network modeling advancing the model's performance. Besides, this also implies that the learnt template brain graphs can be utilized as a plug-in-play manner in brain graph models.

\begin{table}[htbp]
\small
\caption{Results (\%) of TiBGL with different encoders.}
\label{tab:template}
\centering
\begin{tabular*}{0.4\textwidth}{@{\extracolsep{\fill}}ccc}  
\hline
Encoder & ABIDE & CJFH \\
\hline  
MLP  & 65.00$_{\pm 4.31}$  & 75.19$_{\pm 4.96}$ \\
Transformer  & 68.00$_{\pm 5.41}$  & 78.33$_{\pm 3.86}$ \\
CNN & 71.55$_{\pm 2.87}$ & 80.57$_{\pm 2.14}$ \\
\hline
\end{tabular*}
\end{table}

\subsubsection{Effort of Hyper-parameters}
To validate the effort of hyper-parameters $\lambda_1$ and $\lambda_2$ in eq. (\ref{eq:overall}), we conduct the experiments with different settings on the ABIDE dataset. We summarize the experimental result of classification accuracy in Tables \ref{tab:param1}-\ref{tab:param2}. The propose TiBGL can obtain stable results on the ABIDE dataset in most situations. According to results in the Tables \ref{tab:param1}-\ref{tab:param2}, we can readily find that the proposed TaGBL obtains the best performance when $\lambda_1=0.1$ and $\lambda_2=0.005$. More importantly, it's obvious that there exists a wide range for hyper-parameters $\lambda_1$ and $\lambda_2$ in which relatively stable and good results can be readily obtained.

\begin{table}[htbp]
\small
\caption{Results (\%)  with different $\lambda_1$ and $\lambda_2$=0.005.}
\label{tab:param1}
\centering
\begin{tabular*}{0.45\textwidth}{@{\extracolsep{\fill}}ccccc}  
\hline
 $\lambda_1$ & 0.01 & 0.05 & 0.1 & 0.2 \\
 \hline
ACC & 70.52$_{\pm 3.47}$ & 70.23$_{\pm 1.14}$ & 71.55$_{\pm 2.87}$ & 65.57$_{\pm 5.43}$\\
\hline
\end{tabular*}
\end{table}

\begin{table}[htbp]
\small
\caption{Results (\%) with different $\lambda_2$ and $\lambda_1$=0.1.}
\label{tab:param2}
\centering
\begin{tabular*}{0.45\textwidth}{@{\extracolsep{\fill}}ccccc}  
\hline
 $\lambda_2$ & 0.001 & 0.002 & 0.005 & 0.01 \\
 \hline
ACC & 69.88$_{\pm 1.97}$ & 70.44$_{\pm 2.53}$ & 71.55$_{\pm 2.87}$ & 70.32$_{\pm 4.34}$\\
\hline
\end{tabular*}
\end{table}

\subsection{Discussions (Q4)}
Based on the above experimental results and analysis, we can find that the TiBGL not only gets better performance on brain classification tasks but also discovers those insightful connectivity patterns related to brain disorder. For this reason, it's not difficult to observe that the proposed TiBGL has the following advantages in terms of robustness, discrimination, and interpretation of brain network model.
\begin{itemize}
    \item \textbf{Robustness.} TiBGL introduces template brain graphs to overcome the issues of noise and redundant information in original brain graphs, guiding the following tasks to highlight those important brain ROIs as well as their connectivity relationships. The template graphs exploit template graphs from two intra-group and inter-group aspects, which can augment instance-level and group-level brain networks for diagnosing and explaining the neurological disorders. Compared with those works based on dense brain networks, TiBGL adopts one more effective manner to improve its robustness.
    
    \item \textbf{Discrimination.} Compared with current brain network classification works, the redundant and noisy information in the instance-level brain networks has been firstly removed to some extent, augmented by template brain graphs. More importantly, augmented brain networks are relatively easy to be divided into correct groups. Then, CNN model with few parameters is adopted to implement the brain network classification for brain disorders, which is beneficial for end-to-end deep model training with limited subjects. 
    
    \item \textbf{Interpretation.} TiBGL combines the template brain graphs with related neuroscience findings to search meaningful subgraph in brain network, highlighting those important ROI nodes as connectivity patterns. The subgraph can explicitly provide the ROIS sets and their connectivities for disorder analysis, which helps to better understand the neural mechanism of neurological disorders. The results on ABIDE dataset show that the findings of TiBGL can keep coherent with recent neuroscience literatures. 
    
\end{itemize}

\section{Conclusion}\label{conclusion}
Mining human-brain networks to discover patterns that can be used to discriminate between healthy individuals and patients affected by some brain disorders is a fundamental task in neuroscience. To accomplish this target, this paper proposes a novel brain graph learning framework with both discriminative and interpretable abilities, called TiBGL. TiBGL firstly proposes template brain graph learning to extract the template graph for each group, which can be utilized to induce the following tasks of discrimination and interpretation analysis. Notably, the learned template graphs not only highlight those important connectivity patterns in each group but also remove the noise and redundant information in brain networks. Then, template-induced convolutional neural network is designed to fuse rich information from brain graphs and learned template graphs. Moreover, TiBGL provides insightful brain interpretation analysis induced by template graphs to explore meaningful connectivity patterns related to brain disorders. To this end, TiBGL can provide a whole framework with powerful discrimination and interpretation via learnt templates. The experimental results on brain graph datasets can demonstrate that the proposed TiBGL can help to better diagnose and understand the neural mechanism of neurological disorders.

\section*{Acknowledgements}
The authors would like to thank the anonymous reviewers for their insightful comments and suggestions to significantly improve the quality of this paper. 

\bibliographystyle{IEEEtran}
\bibliography{IEEEexample}

\end{document}